 \documentclass[tablecaption=bottom,wcp]{jmlr} 

\usepackage{booktabs}
\usepackage[load-configurations=version-1]{siunitx} 

\theorembodyfont{\upshape}
\theoremheaderfont{\scshape}
\theorempostheader{:}
\theoremsep{\newline}

\jmlrproceedings{AABI 2021}{3rd Symposium on Advances in Approximate Bayesian Inference, 2021}

\title[Gradient-Free Adversarial Attacks for BNNs]{Gradient-Free Adversarial Attacks for\\ Bayesian Neural Networks}

 \author{\Name{Matthew Yuan} \addr Princeton University \Email{my4@princeton.edu}\\
  \Name{Matthew Wicker} \addr University of Oxford \Email{matthew.wicker@cs.ox.ac.uk}\\
  \Name{Luca Laurenti} \addr University of Oxford \Email{luca.laurenti@cs.ox.ac.uk}\\
  }

\usepackage{amsmath}
\usepackage{amssymb}
\usepackage{mathtools}
\usepackage{algorithm}
\usepackage{algpseudocode}

\newcommand{\defeq}{\coloneqq}

\DeclareMathOperator{\fitness}{Fitness}
\DeclareMathOperator{\select}{Select}
\DeclareMathOperator{\cross}{Cross}
\DeclareMathOperator{\mutate}{Mutate}
\DeclareMathOperator{\size}{size}

\DeclareMathOperator{\clip}{clip}
\DeclareMathOperator{\argmax}{arg\,max}
\renewcommand{\vec}[1]{\mathbf{#1}}

\begin{document}

\maketitle

\begin{abstract}
The existence of adversarial examples, i.e., small perturbations to their inputs that can cause a misclassification, underscores the importance of understanding the robustness of machine learning models. Bayesian neural networks (BNNs), due to their calibrated uncertainty,  have been shown to posses favorable adversarial robustness properties. However, when approximate Bayesian inference methods are employed, the adversarial robustness of BNNs is still not well understood.  In this work, we employ gradient-free optimization methods in order to find adversarial examples for BNNs. In particular, we consider genetic algorithms, surrogate models, as well as zeroth order optimization methods and adapt them to the goal of finding adversarial examples for BNNs.  In an empirical evaluation on the MNIST and Fashion MNIST datasets, we show that for various approximate Bayesian inference methods the usage of  gradient-free algorithms
can greatly improve the rate of finding adversarial examples compared to state-of-the-art gradient-based methods. 
\end{abstract}


\section{Introduction}
\label{sec:intro}

Deep Neural Networks (NN) have shown state-of-the-art performance on many tasks, including image recognition \citep{krizhevsky2017imagenet} and reinforcement learning \citep{schulman2015trust}. However, the vulnerability of NNs to  adversarial examples questions their applicability to safety-critical scenarios, where a failure of a learning model can have catastrophic consequences \citep{goodfellow2014explaining}. Because of their principled treatment of uncertainty, Bayesian Neural Networks (BNNs), i.e., neural networks with a prior distribution placed over their parameters, have been proposed as a more robust learning model compared to standard SGD-trained NNs \citep{carbone,bekasov2018bayesian}.  Nevertheless, the theoretical results are all limited to the idealised case where the posterior distribution of a BNN is computed exactly via the Bayes' rule. In practice, exact computation of the posterior is infeasible and approximate inference methods with a finite amount of data are employed \citep{neal2012bayesian}. As a result, it is still unclear how to best find  adversarial examples for commonly employed approximate Bayesian inference methods. 

In this paper, we start from the observation that for fully trained BNNs the gradient of the loss tends to be uninformative \citep{carbone}, and we adapt  gradient-free optimization methods for finding adversarial attacks on BNNs. 
In particular, we consider: zeroth order optimization \citep{chen2017zoo}, optimization with surrogate gradients \citep{athalye2018obfuscated}, and genetic algorithms \citep{alzantot2019genattack}. We perform an empirical evaluation of various different NN architectures on both the MNIST and FashionMNIST datasets for networks trained with five different commonly employed approximate inference methods for BNNs. 
We find that the presented algorithms can offer significant improvements, finding adversarial examples for up to 40\% of images which remained correctly classified when using state-of-the-art, first-order attacks. \footnote{All source code to reproduce these results including the posteriors used and BNN training scripts can be found at: https://github.com/matthewwicker/GradientFreeAttacksBNNs}

\section{Adversarial Attacks for BNNs}
\label{sec:AdvAttacks}
Bayesian modelling aims to capture the intrinsic uncertainty of data driven models. 
Consider a classification problem with $n_C$ classes for a neural network $f^{\mathbf{w}}(x)$ with input $x \in \mathcal{R}^{m}$ and network parameters (weights and biases) $\mathbf{w}$, then  one starts with a prior distribution over the network parameters $p(\mathbf{w})$. The fit of the  weights $\mathbf{w}$ to the data $D$ is computed through the likelihood $p(D\vert\mathbf{w})$. Bayesian inference combines likelihood and prior via the Bayes theorem to obtain a {\it posterior} distribution $ p\left(\mathbf{w}\vert D\right)\propto  p\left(D\vert\mathbf{w}\right)p\left(\mathbf{w}\right)$.

Computing the posterior distribution $p\left(\mathbf{w}\vert D\right)$ exactly is not possible in general for NNs. Asymptotically exact samples from $p\left(\mathbf{w}\vert D\right)$ can be obtained via approximate inference methods such as Hamiltonian Monte Carlo (HMC) \citep{neal2011mcmc}, while approximate samples can be obtained more cheaply via Variational Inference (VI) \citep{blundell2015weight}. 
Independently of the methods employed to compute the posterior distribution, predictions at a new input $x^*$ are obtained from an ensemble of $n$ NNs, each with its individual weights drawn from the approximate posterior distribution $p(\mathbf{w}|D):$\begin{equation}\begin{split}
   &\langle f^{\mathbf{w}}(x^*)\rangle_{p\left(\mathbf{w}\vert D\right)}
    \simeq \frac{1}{n}\sum_{i=1}^n f^{\mathbf{w}_i}(x^*)\qquad \mathbf{w}_i\sim p\left(\mathbf{w}\vert D\right)
\end{split}\label{predDist}\end{equation} 
where $\langle\cdot\rangle_p$ denotes expectation w.r.t.\ the distribution $p$. Eqn \eqref{predDist}  is the expectation of the {\it predictive distribution} of the BNN. 

Given an input point $x^*$ and a strength (i.e.\ maximum perturbation magnitude) $\epsilon>0$,
an adversarial example is a point $\tilde{x}$  such that $|x^*-\tilde{x}|\leq \epsilon $ and
$$ \text{argmax}_{i\in \{1,...,n_C \}} \langle f^{\mathbf{w}}_i(x^*)\rangle_{p\left(\mathbf{w}\vert D\right)} \neq  \text{argmax}_{i\in \{1,...,n_C \}} \langle f^{\mathbf{w}}_i(\tilde{x})\rangle_{p\left(\mathbf{w}\vert D\right)},$$
where $f^{\mathbf{w}}_i$ is the i-th component of $f^{\mathbf{w}}$, and $\vert\cdot\vert$ is a suitable similarity metric (taken here to be the $l_\infty$ norm). As $f^{\mathbf{w}}$ is non-linear, solving the above optimization problem exactly is infeasible and several approximate solution methods have been proposed. Among them gradient-based attacks are arguably the most prominent, efficient, and effective.
In particular, these methods, which include FGSM \citep{goodfellow2014explaining} and PGD \citep{pgdattack},  which rely on the gradient of the loss $L(x,\mathbf{w}_i)$ function wrt the input point $x$.
However, it has been recently shown in \citep{carbone} that for overpametrised BNNs in the limit of high accuracy, exact inference, and infinite amount of data it holds that:
\begin{equation}
    \label{Eqn:ZeroGradient}
    \langle\nabla_{\mathbf{x}}L(\mathbf{x},\mathbf{w})\rangle_{p(\mathbf{w}|D)}=0. 
\end{equation} 
Such a vanishing behavior on the gradient (which is known to be not true for SGD-trained NNs) questions the efficacy of gradient-based attacks for BNNs.  In what follows, we employ several optimization methods which do not directly rely on the gradient of the loss to attack BNNs and empirically evaluate them in Section \ref{sec:Experimets}. 

\subsection{Attacking BNNs without First-Order Information}
Attacking deterministic neural networks when the gradient information is missing, unavailable, or uncomputable has been studied in prior works, see e.g., \citep{papernot2017practical,athalye2018obfuscated,alzantot2019genattack}. Whereas several works have studied the robustness of BNNs \citep{cardelli2019robustness, wicker2020probabilistic, michelmore2019uncertainty} and developed more robust training methods \citep{liu2018adv}, attacking a BNN when the gradient is uninformative or unavailable has not been considered in detail yet. Below, we adapt some of the more commonly gradient-free optimization approaches for this setting.

\subsubsection{Zeroth Order Adversarial Attacks}

As a first approach we consider numerical approximations of the gradient via Zeroth order optimization (ZOO) methods \citep{chen2017zoo, ilyas2018black, wicker2018feature}. We consider these methods because it has been shown that for NNs with obfuscated gradients (i.e. when a NN is purposefully made non-differentiable or their gradients vanish as for BNNs) numerical approximations of the gradient  can still offer useful information for finding adversarial attacks  \citep{athalye2018obfuscated, papernot2017practical}. In particular, these methods can be adapted to BNNs by making various queries to the predictive distribution to compute a finite difference approximation of the gradient of the expectation of the loss. The resulting approximated gradient is then feed  into standard gradient-based algorithms, such as the fast gradient sign method (FGSM) \citep{goodfellow2014explaining} and projected gradient descent (PGD) \citep{pgdattack}.   

\subsubsection{Backwards Pass Differentiable Approximations (BPDA)}

Backwards Pass Differentiable Approximation (BPDA) attacks aim to learn a differentiable surrogate model in order to mimic the decision boundary of the classifier under consideration \citep{papernot2017practical, athalye2018obfuscated}. The BPDA attack can be easily adapted to BNNs by learning a SGD-trained NN on a dataset where the label of each image in the training set is replaced by the predictive distribution of the BNN on that image. Once this new NN has been learned, one can run standard gradient-based attacks on this network.

\subsubsection{Genetic Algorithm (GA) for Adversarial Attacks}


Both ZOO and BPDA involve approximating the gradient of the BNN wrt a given input. This may be limiting for BNNs (see Eqn \eqref{Eqn:ZeroGradient}). In contrast, in this subsection, we consider a genetic algorithm similar to that proposed in \citep{alzantot2019genattack} which proceeds in a completely gradient-free manner.  Given an input point $x^*$ the genetic algorithm attack (GA) proceeds by first sampling $k$ vectors in an $\epsilon-$ball around $x^*$ from a uniform distribution. These vectors are then each added to the original input to create a set of $k$ candidate adversarial examples. For each candidate, the BNN predictive distribution is queried and each of the $k$ predictions are compared with the original BNN prediction and are scored based on a fitness function. In our case, this fitness function is taken to be the maximizing the loss function (see Eqn~\ref{eqn:fitness} in the Appendix). Tournament selection is then used to decide which modifications to keep and which to discard, that is, we randomly draw (with replacement) $j$ pairs of members from the population and select to keep the member with higher fitness. New members are made by crossing the winners from the tournament selection and some of the resulting set are randomly mutated before repeating these steps. After repeating this process $m$ times (each called a generation), the modification with the highest fitness is returned as the adversarial perturbation. A further description and pseudocode are given in the appendix.

\begin{figure*}[ht]
\centering
\includegraphics[width=0.700\textwidth]{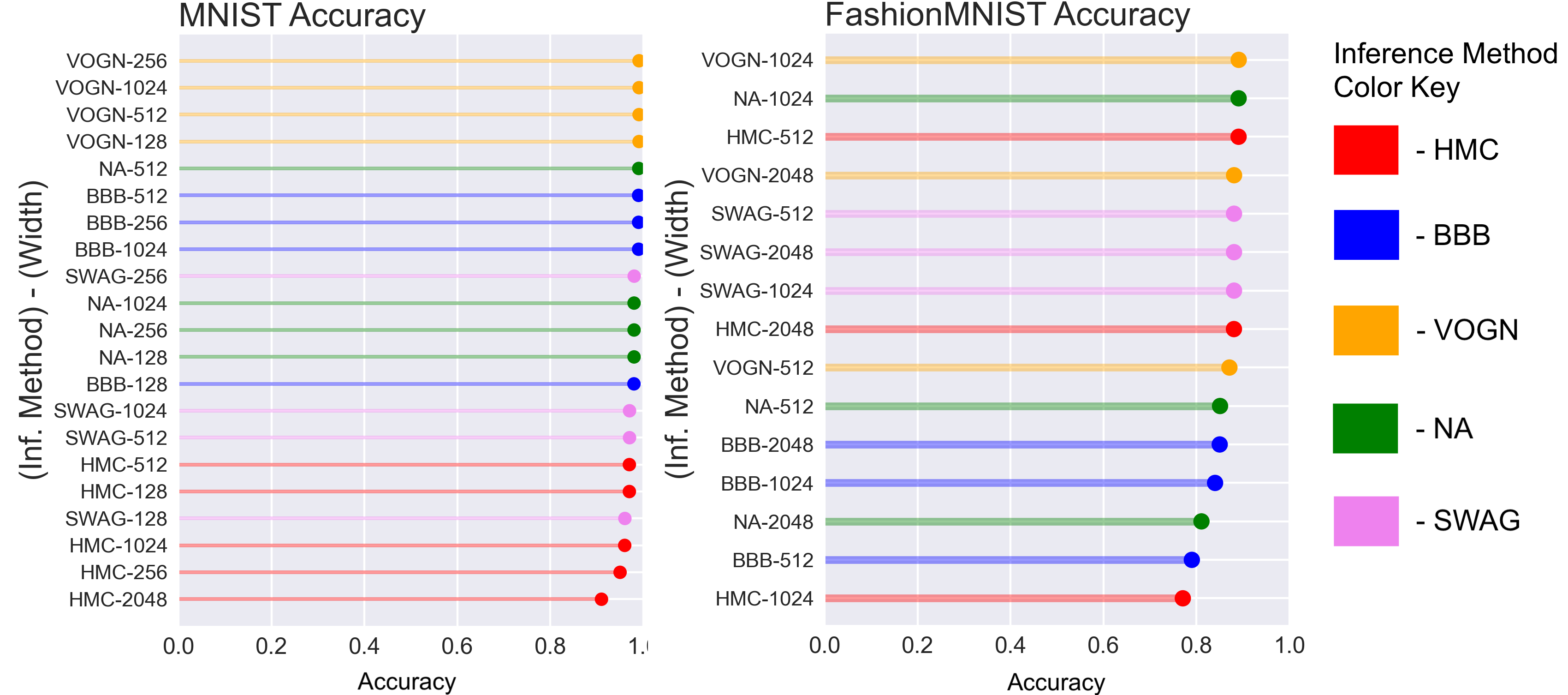}
\caption{Each point in the graph represents the accuracy of the predictive distribution for approximate posterior. Each value is estimated from 500 test set images. 
\textbf{Left Column:} MNIST test set accuracy is $\geq$95\% for each combination of training method and architecture
\textbf{Centre Column:} FashionMNIST test set accuracy indicates that the approximate BNNs achieve between 80\% and 90\% accuracy.
\textbf{Right Column:} Color coordinated key for each inference method that will be used throughout each figure.}\label{fig:accuracies}
\end{figure*}

\begin{figure*}[ht]
\centering
\includegraphics[width=1.1\textwidth]{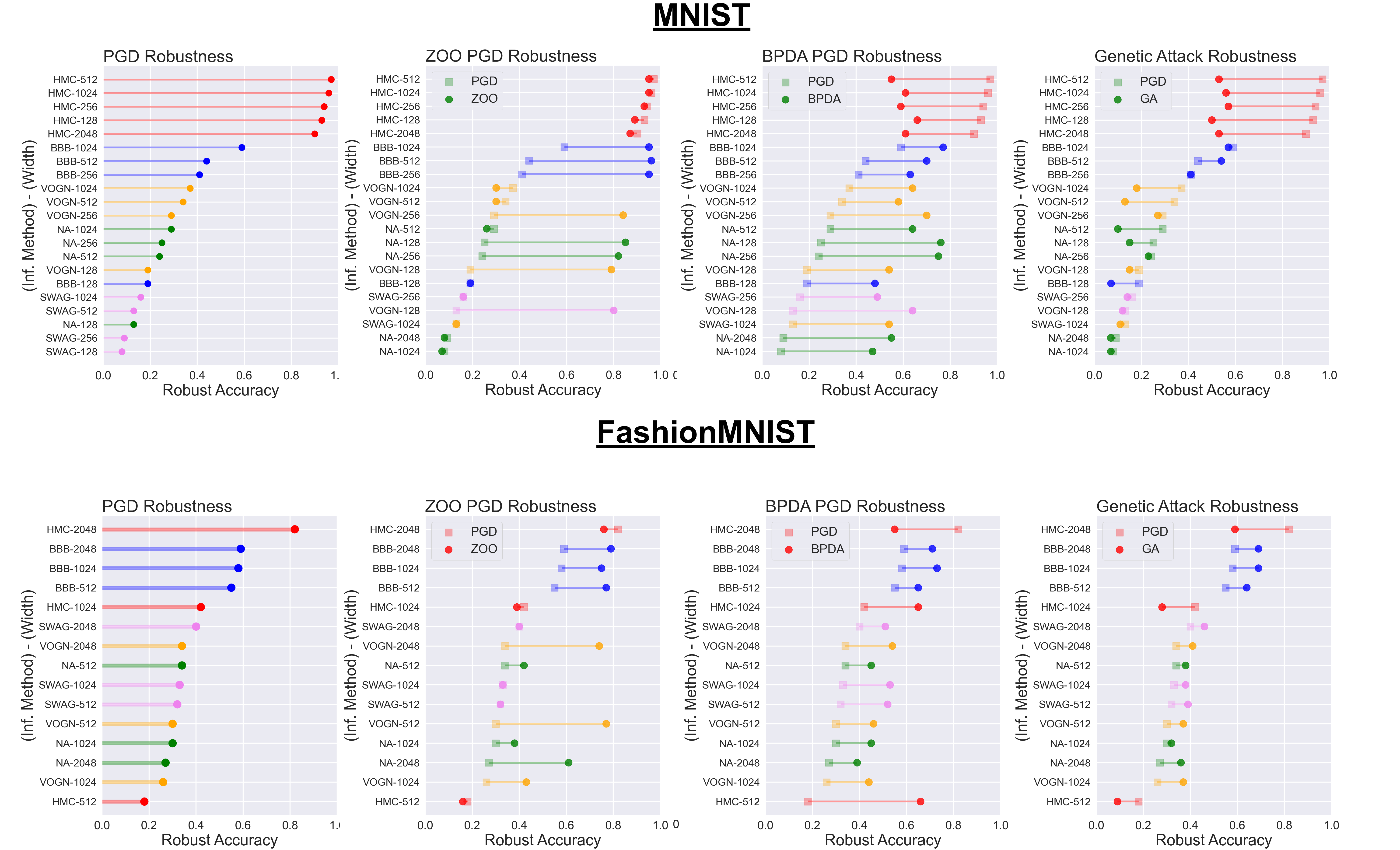}
\vspace{-1em}
\caption{
Robust Accuracy of each posterior wrt different attack algorithms. On the far left we plot as dots the robust accuracy of posteriors against PGD. For all other attack algorithms, in order to compare directly with PGD, we plot the PGD performance as a square and draw a line to the robust accuracy of the posterior against that particular attack (given in the title). 
\textbf{Top Row, left to right:} Robust Accuracy on the MNIST dataset for each approximate Bayesian inference method against PGD, PGD with ZOO, PGD with BPDA, and GA. We observe that in all cases GA outperforms other methods.
\textbf{Top Row, left to right:} Robust Accuracy on the FashionMNIST dataset for each approximate Bayesian inference method against PGD, PGD with ZOO, PGD with BPDA, and GA. We observe that GA outperforms other attacks for HMC, while for other training methods PGD obtains the best performances.
}\label{fig:mnist}
\end{figure*}

\section{Empirical Results}
\label{sec:Experimets}

We conduct an empirical evaluation of the proposed methods against state-of-the-art gradient-based methods (PGD given here and FGSM in the appendix) for various NN architectures trained with different approximate inference methods, including Hamiltonian Monte Carlo (HMC, \cite{neal2011mcmc}), Bayes by Backprop (BBB, \cite{blundell2015weight}), Variational Online Guass Newton (VOGN, \cite{khan2018fast}), NoisyAdam (NA, \cite{zhang2018noisy}), and Stochastic Weight Averaging - Gaussian (SWAG, \cite{maddox}). For each of the BNNs discussed, we select the prior distribution based on the Glorot normal distribution \citep{sutskever2013importance}.
For MNIST we fix the strength of the adversarial perturbation $\epsilon = 0.1$ and for FashionMNIST we instead consider $\epsilon = 0.05$. For each PGD attack we consider 5 iterations with 1 restart. We note that each iteration of PGD requires estimation of the expectation of the gradient via Monte Carlo integration making this setting as computationally expensive as 1000 iterations of PGD in the deterministic case (assuming 100 steps for the Monte Carlo integration).

We perform our experiments on the MNIST and Fashion MNIST data-sets. We note that in our evaluation, we do not consider bigger and more complex datasets, such as CIFAR-10, because we could not train HMC and BBB on those datasets with good accuracy.

\paragraph{Evaluation of Robust Accuracy on MNIST}
We start our evaluation with the MNIST dataset. In Figure \ref{fig:accuracies} we report the accuracy of the various NNs on clean data. For this evaluation we consider networks with a single hidden layer of varying width (from 128 to 2048) and we observe that for all the models we obtain an accuracy  $\geq 95\%$.

In Figure \ref{fig:mnist} (top row) we report the computed robustness to adversarial attacks  for all the various networks and methods. The robustness is measured with the `Robust Accuracy', which is a standard measure of robustness \citep{szegedy2013intriguing} defined as the percentage of test points that remain successfully classified after a given attack (i.e., the smaller the Robust Accuracy, the less robust the network).
From Eqn \eqref{Eqn:ZeroGradient} we expect that, at least for HMC, which is known to produce asymptotically exact samples from the posterior,  all the networks are robust to gradient-based attacks. This is indeed the case. However,  GA which does not rely on the gradient, is up to $40\%$ more effective in these cases. Note that the obtained values of robustness are still considerably higher than those commonly reported for SGD-trained NNs \citep{carbone}. 

For other approximate inference methods, GA is still more effective than PGD. However, we observe that the more approximated is the posterior computation, the more effective are gradient-based attacks. In fact, for all SWAG-trained networks (which in the degenerate case is exactly SGD-training), the robust accuracy to PGD drops from the $\sim 95\%$ of HMC  to $\sim 10\%$ (i.e. random guessing). 
Finally, we observe that, in the case of HMC, ZOO performs comparably with PGD,
while BPDA is considerably better and PGD, even if less effective than EA. This is due to the fact that at no point during the optimization does BPDA query the posterior predictive distribution of the BNN under consideration. This again confirms that for BNNs finding adversarial attacks by relying on the loss gradient can be misleading.

\paragraph{Evaluation of Robust Accuracy on Fashion MNIST}
We continue our evaluation with the Fashion MNIST dataset. In Figure \ref{fig:accuracies} (centre column) we report the accuracy of the various trained NNs. For this evaluation we consider networks with $2$ hidden layers with same number of neurons for each layer and of varying width (512, 1024, and 2048). 

In Figure \ref{fig:mnist} (bottom row) we report the robust accuracy  for all the various networks and methods. We observe that HMC and BBB trained networks are able to achieve notable success in resisting adversarial examples. Similar to what is observed in MNIST, we see that for HMC trained networks, GA is the most effective algorithm, showing up to 30\% improvements over PGD attacks. This again confirms how for BNNs gradient-based attacks can be surprisingly ineffective compared to black-box optimization methods. 
However, when we move to other approximate inference methods we see that these offer less protection against gradient-based attacks. In fact, in these cases PGD outperforms all the other methods with GA obtaining similar, but slightly worse, results. As already noticed in \citep{carbone, gal2018sufficient}, this may be due to the fact that Variational Inference (VI) methods, for larger and more complex datasets, may tend to poorly approximate the uncertainty and lead to a posterior that is far from the true distribution \citep{cardelli2019statistical}. As a consequence, the loss gradient may become meaningful and the resulting networks fragile to gradient-based attacks such as PGD.

\section{Conclusion}
We investigated gradient-free optimization algorithms for finding adversarial examples on BNNs.  In an empirical evaluation on the MNIST and FashionMNIST datasets, we showed that gradient-free algorithms can be more effective than gradient-based methods, especially for NNs trained with approximate inference methods that compute a more accurate approximation of the true posterior.

\bibliography{jmlr-sample}

\begin{thebibliography}{29}
\providecommand{\natexlab}[1]{#1}
\providecommand{\url}[1]{\texttt{#1}}
\expandafter\ifx\csname urlstyle\endcsname\relax
  \providecommand{\doi}[1]{doi: #1}\else
  \providecommand{\doi}{doi: \begingroup \urlstyle{rm}\Url}\fi

\bibitem[Alzantot et~al.(2019)Alzantot, Sharma, Chakraborty, Zhang, Hsieh, and
  Srivastava]{alzantot2019genattack}
Moustafa Alzantot, Yash Sharma, Supriyo Chakraborty, Huan Zhang, Cho-Jui Hsieh,
  and Mani~B Srivastava.
\newblock Genattack: Practical black-box attacks with gradient-free
  optimization.
\newblock In \emph{Proceedings of the Genetic and Evolutionary Computation
  Conference}, pages 1111--1119, 2019.

\bibitem[Athalye et~al.(2018)Athalye, Carlini, and
  Wagner]{athalye2018obfuscated}
Anish Athalye, Nicholas Carlini, and David Wagner.
\newblock Obfuscated gradients give a false sense of security: Circumventing
  defenses to adversarial examples.
\newblock \emph{arXiv preprint arXiv:1802.00420}, 2018.

\bibitem[Bekasov and Murray(2018)]{bekasov2018bayesian}
Artur Bekasov and Iain Murray.
\newblock Bayesian adversarial spheres: Bayesian inference and adversarial
  examples in a noiseless setting.
\newblock \emph{arXiv preprint arXiv:1811.12335}, 2018.

\bibitem[Blundell et~al.(2015)Blundell, Cornebise, Kavukcuoglu, and
  Wierstra]{blundell2015weight}
Charles Blundell, Julien Cornebise, Koray Kavukcuoglu, and Daan Wierstra.
\newblock Weight uncertainty in neural networks.
\newblock \emph{arXiv preprint arXiv:1505.05424}, 2015.

\bibitem[Carbone et~al.(2020)Carbone, Wicker, Laurenti, Patane, Bortolussi, and
  Sanguinetti]{carbone}
G.~Carbone, M.~Wicker, L.~Laurenti, A.~Patane, L.~Bortolussi, and
  G.~Sanguinetti.
\newblock Robustness of bayesian neural networks to gradient-based attacks.
\newblock \emph{NeurIPS}, 2020.

\bibitem[Cardelli et~al.(2019{\natexlab{a}})Cardelli, Kwiatkowska, Laurenti,
  Paoletti, Patane, and Wicker]{cardelli2019statistical}
Luca Cardelli, Marta Kwiatkowska, Luca Laurenti, Nicola Paoletti, Andrea
  Patane, and Matthew Wicker.
\newblock Statistical guarantees for the robustness of bayesian neural
  networks.
\newblock \emph{IJCAI}, 2019{\natexlab{a}}.

\bibitem[Cardelli et~al.(2019{\natexlab{b}})Cardelli, Kwiatkowska, Laurenti,
  and Patane]{cardelli2019robustness}
Luca Cardelli, Marta Kwiatkowska, Luca Laurenti, and Andrea Patane.
\newblock Robustness guarantees for bayesian inference with gaussian processes.
\newblock In \emph{Proceedings of the AAAI Conference on Artificial
  Intelligence}, volume~33, pages 7759--7768, 2019{\natexlab{b}}.

\bibitem[Chen et~al.(2017)Chen, Zhang, Sharma, Yi, and Hsieh]{chen2017zoo}
Pin-Yu Chen, Huan Zhang, Yash Sharma, Jinfeng Yi, and Cho-Jui Hsieh.
\newblock Zoo: Zeroth order optimization based black-box attacks to deep neural
  networks without training substitute models.
\newblock In \emph{Proceedings of the 10th ACM Workshop on Artificial
  Intelligence and Security}, pages 15--26, 2017.

\bibitem[Gal and Smith(2018)]{gal2018sufficient}
Yarin Gal and Lewis Smith.
\newblock Sufficient conditions for idealised models to have no adversarial
  examples: a theoretical and empirical study with bayesian neural networks.
\newblock \emph{arXiv preprint arXiv:1806.00667}, 2018.

\bibitem[Goodfellow et~al.(2014)Goodfellow, Shlens, and
  Szegedy]{goodfellow2014explaining}
Ian~J Goodfellow, Jonathon Shlens, and Christian Szegedy.
\newblock Explaining and harnessing adversarial examples.
\newblock \emph{arXiv preprint arXiv:1412.6572}, 2014.

\bibitem[Ilyas et~al.(2018)Ilyas, Engstrom, Athalye, and Lin]{ilyas2018black}
Andrew Ilyas, Logan Engstrom, Anish Athalye, and Jessy Lin.
\newblock Black-box adversarial attacks with limited queries and information.
\newblock \emph{arXiv preprint arXiv:1804.08598}, 2018.

\bibitem[Khan et~al.(2018)Khan, Nielsen, Tangkaratt, Lin, Gal, and
  Srivastava]{khan2018fast}
Mohammad~Emtiyaz Khan, Didrik Nielsen, Voot Tangkaratt, Wu~Lin, Yarin Gal, and
  Akash Srivastava.
\newblock Fast and scalable bayesian deep learning by weight-perturbation in
  adam.
\newblock \emph{arXiv preprint arXiv:1806.04854}, 2018.

\bibitem[Krizhevsky et~al.(2017)Krizhevsky, Sutskever, and
  Hinton]{krizhevsky2017imagenet}
Alex Krizhevsky, Ilya Sutskever, and Geoffrey~E Hinton.
\newblock Imagenet classification with deep convolutional neural networks.
\newblock \emph{Communications of the ACM}, 60\penalty0 (6):\penalty0 84--90,
  2017.

\bibitem[Liu et~al.(2018)Liu, Li, Wu, and Hsieh]{liu2018adv}
Xuanqing Liu, Yao Li, Chongruo Wu, and Cho-Jui Hsieh.
\newblock Adv-bnn: Improved adversarial defense through robust bayesian neural
  network.
\newblock \emph{arXiv preprint arXiv:1810.01279}, 2018.

\bibitem[Maddox et~al.(2019)Maddox, Izmailov, Garipov, Vetrov, and
  Wilson]{maddox}
Wesley~J Maddox, Pavel Izmailov, Timur Garipov, Dmitry~P Vetrov, and
  Andrew~Gordon Wilson.
\newblock A simple baseline for bayesian uncertainty in deep learning.
\newblock \emph{Advances in Neural Information Processing Systems},
  32:\penalty0 13153--13164, 2019.

\bibitem[Madry et~al.(2018)Madry, Makelov, Schmidt, Tsipras, and
  Vladu]{pgdattack}
Aleksander Madry, Aleksandar Makelov, Ludwig Schmidt, Dimitris Tsipras, and
  Adrian Vladu.
\newblock Towards deep learning models resistant to adversarial attacks.
\newblock In \emph{International Conference on Learning Representations}, 2018.

\bibitem[Michelmore et~al.(2020)Michelmore, Wicker, Laurenti, Cardelli, Gal,
  and Kwiatkowska]{michelmore2019uncertainty}
Rhiannon Michelmore, Matthew Wicker, Luca Laurenti, Luca Cardelli, Yarin Gal,
  and Marta Kwiatkowska.
\newblock Uncertainty quantification with statistical guarantees in end-to-end
  autonomous driving control.
\newblock \emph{2020 IEEE International Conference on Robotics and Automation
  (ICRA)}, pages 7344--7350, 2020.

\bibitem[Neal(2012)]{neal2012bayesian}
Radford~M Neal.
\newblock \emph{Bayesian learning for neural networks}, volume 118.
\newblock Springer Science \& Business Media, 2012.

\bibitem[Neal et~al.(2011)]{neal2011mcmc}
Radford~M Neal et~al.
\newblock Mcmc using hamiltonian dynamics.
\newblock \emph{Handbook of markov chain monte carlo}, 2\penalty0
  (11):\penalty0 2, 2011.

\bibitem[Papernot et~al.(2017)Papernot, McDaniel, Goodfellow, Jha, Celik, and
  Swami]{papernot2017practical}
Nicolas Papernot, Patrick McDaniel, Ian Goodfellow, Somesh Jha, Z~Berkay Celik,
  and Ananthram Swami.
\newblock Practical black-box attacks against machine learning.
\newblock In \emph{Proceedings of the 2017 ACM on Asia conference on computer
  and communications security}, pages 506--519, 2017.

\bibitem[Schulman et~al.(2015)Schulman, Levine, Abbeel, Jordan, and
  Moritz]{schulman2015trust}
John Schulman, Sergey Levine, Pieter Abbeel, Michael Jordan, and Philipp
  Moritz.
\newblock Trust region policy optimization.
\newblock In \emph{International conference on machine learning}, pages
  1889--1897, 2015.

\bibitem[Soni and Kumar(2014)]{mutation}
Nitasha Soni and Tapas Kumar.
\newblock Study of various mutation operators in genetic algorithms, 2014.

\bibitem[Sutskever et~al.(2013)Sutskever, Martens, Dahl, and
  Hinton]{sutskever2013importance}
Ilya Sutskever, James Martens, George Dahl, and Geoffrey Hinton.
\newblock On the importance of initialization and momentum in deep learning.
\newblock In \emph{International conference on machine learning}, pages
  1139--1147, 2013.

\bibitem[Szegedy et~al.(2013)Szegedy, Zaremba, Sutskever, Bruna, Erhan,
  Goodfellow, and Fergus]{szegedy2013intriguing}
Christian Szegedy, Wojciech Zaremba, Ilya Sutskever, Joan Bruna, Dumitru Erhan,
  Ian Goodfellow, and Rob Fergus.
\newblock Intriguing properties of neural networks.
\newblock \emph{arXiv preprint arXiv:1312.6199}, 2013.

\bibitem[Umbarkar and Sheth(2015)]{crossover}
Dr.~Anantkumar Umbarkar and P.~Sheth.
\newblock Crossover operators in genetic algorithms: A review.
\newblock \emph{ICTACT Journal on Soft Computing (Volume: 6 , Issue: 1)}, 6, 10
  2015.
\newblock \doi{10.21917/ijsc.2015.0150}.

\bibitem[Wicker et~al.(2018)Wicker, Huang, and Kwiatkowska]{wicker2018feature}
Matthew Wicker, Xiaowei Huang, and Marta Kwiatkowska.
\newblock Feature-guided black-box safety testing of deep neural networks.
\newblock In \emph{International Conference on Tools and Algorithms for the
  Construction and Analysis of Systems}, pages 408--426. Springer, 2018.

\bibitem[Wicker et~al.(2020)Wicker, Laurenti, Patane, and
  Kwiatkowska]{wicker2020probabilistic}
Matthew Wicker, Luca Laurenti, Andrea Patane, and Marta Kwiatkowska.
\newblock Probabilistic safety for bayesian neural networks.
\newblock \emph{UAI}, 2020.

\bibitem[Zhang et~al.(2018)Zhang, Sun, Duvenaud, and Grosse]{zhang2018noisy}
Guodong Zhang, Shengyang Sun, David Duvenaud, and Roger Grosse.
\newblock Noisy natural gradient as variational inference.
\newblock In \emph{International Conference on Machine Learning}, pages
  5852--5861, 2018.

\bibitem[Zhong et~al.()Zhong, Hu, Gu, and Zhang]{selection}
Jinghui Zhong, Xiaomin Hu, Min Gu, and Jun Zhang.
\newblock Comparison of performance between different selection strategies on
  simple genetic algorithms.

\end{thebibliography}

\appendix
\section{FGSM Results}

\begin{figure*}[ht]
\centering
\includegraphics[width=1.0\textwidth]{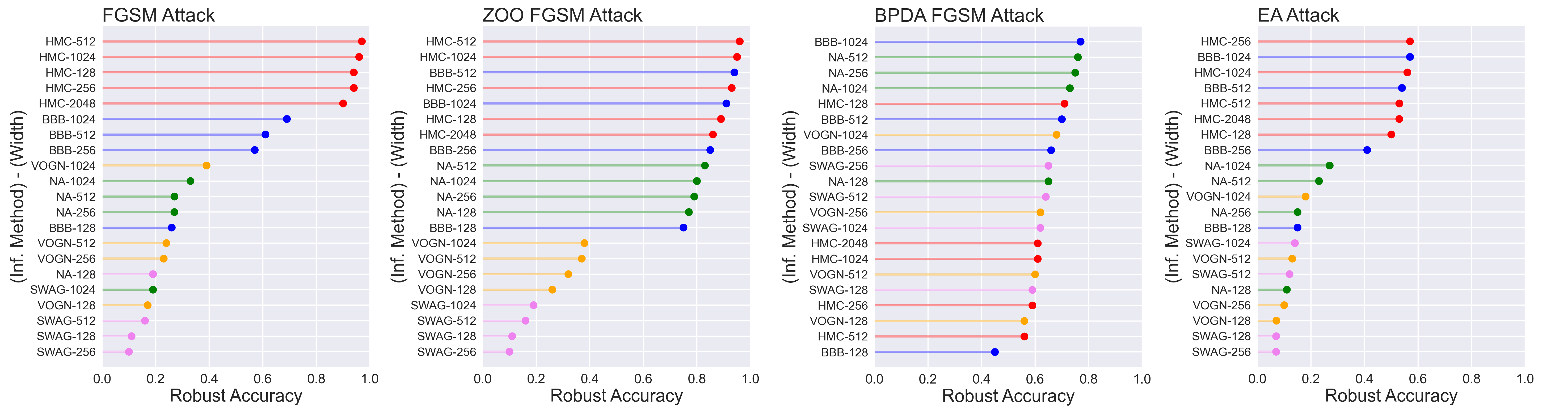}

\caption{
Robust Accuracy of each posterior wrt different attack algorithms. On the far left we plot as dots the robust accuracy of posteriors against FGSM. 
\textbf{Left to Right:} Robust Accuracy on the MNIST dataset for each approximate Bayesian inference method against FGSM, FGSM with ZOO approximate gradients, FGSM with BPDA approximate gradients, GA. We observe that in all cases GA outperforms other methods.
}\label{fig:fgsm}
\end{figure*}

In Figure~\ref{fig:fgsm} we report the attack results for the investigated attacks on the studied posteriors with FGSM.

\section{Genetic Attack Details}

The Genetic Attack (GA) follows the conventional structure of a genetic algorithm:
\begin{enumerate}
    \itemsep0em
    \item Initialize a population,
    \item compute the fitness of each member of the current population,
    \item (selection) select certain members of the current population to be parents for the next population,
    \item (crossover) cross the selected parents to form children,
    \item (mutation) randomly modify the children and collect them to form the next population, then
    \item repeat steps 2–4 for multiple generations until a stopping condition is reached.
\end{enumerate}
The goal of a genetic algorithm is to increase the overall fitness of the population. In our case, the population consists of a collection of $N$ perturbation vectors which we may add to an image $x$ to turn it into an adversarial image. We define the fitness at class $c$ of a member $m$ of the population to be the negative logarithm of the $c$th coordinate of the BNN's prediction for $x + m$ after clipping to make every coordinate in $[0, 1]$.
\begin{equation} \label{eqn:fitness}
\fitness(m, c) \defeq -\ln\big((\langle{f^{\vec{w}}}(\clip(x + m))\rangle_{p(\vec{w}|D)})_c \big),
\end{equation}

The fitness of $m$ at thew correct class measures how adversarial is. If the $\fitness(m, c)$ is high, then the BNN predicts that $x + m$ belongs to class $c$ with low probability.

GeneticAttack initializes the population as $N$ random $\{-\varepsilon, \varepsilon\}$-valued vectors with the same dimensions as the input space. Such vectors are the largest perturbations allowed by our constraint, so intuitively they have the best chance at having a high fitness.

Algorithm \ref{alg:select} is our selection operator. It is a standard operator in genetic algorithms called tournament selection \cite{selection}, and we use a tournament size of two. We randomly draw (with replacement) $N$ pairs of members from the population and select from each pair the member with a higher fitness.

\begin{figure}
    \centering
    \includegraphics[width=0.3\textwidth]{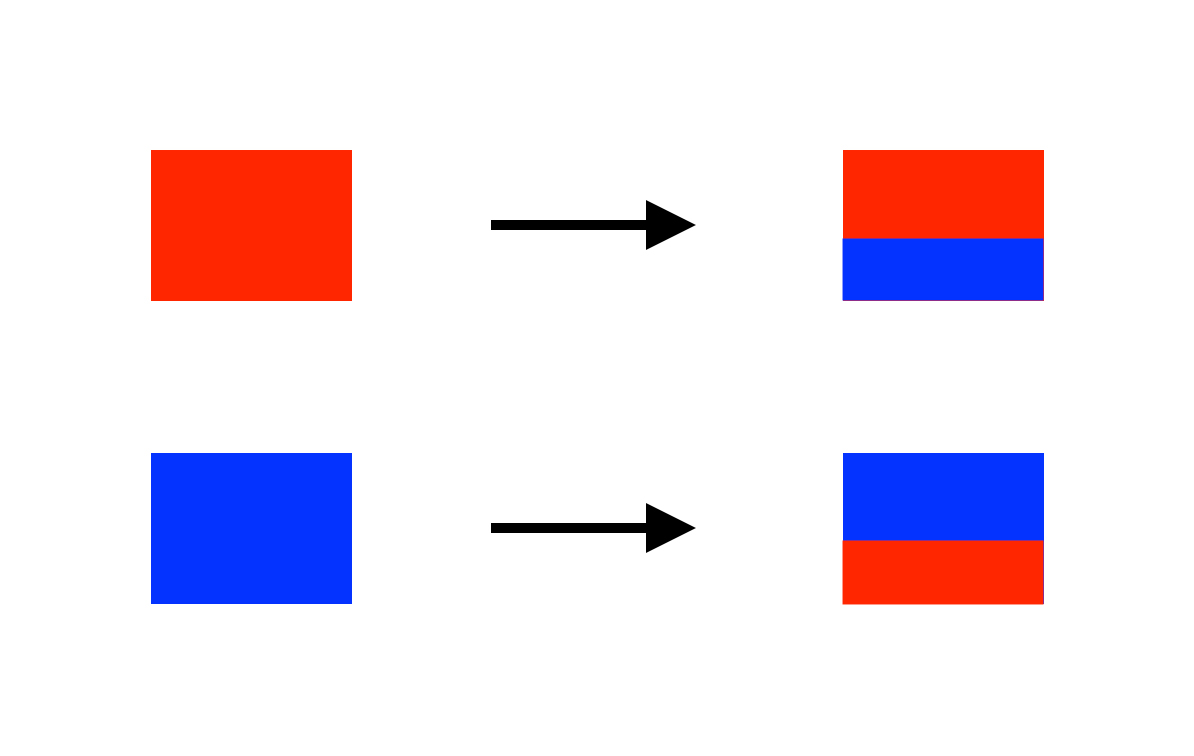}
    \caption{Horizontal crossover.}
    \label{fig:cross}
\end{figure}

Algorithm \ref{alg:cross} is our crossover operation. It is an adaptation of the standard single-point crossover \cite{crossover} to two-dimensional inputs. Given two parent images, we randomly select a location to cross the parents horizontally or vertically. Figure \ref{fig:cross} illustrates horizontal crossover.

Algorithm \ref{alg:mutate} is our mutation operator. It decides to mutate each child with probability $R$, and mutation consists of flipping the sign on precisely one randomly chosen pixel of that child. Our mutation operator might be considered a sparse form of flip mutation \cite{mutation}. In the case of MNIST and Fashion MNIST, we are flipping single coordinates. In the case of CIFAR10 we are flipping a triple corresponding to perturbations in a pixel's RGB values.

GeneticAttack stops after $G$ iterations, and it returns the original image $x$ plus the population member with the highest fitness.

\begin{algorithm}[H] \footnotesize
\label{alg:full}
\SetAlgoLined
\KwData{BNN, original image $x$, correct class $c$, number of generations $G$, population size $N$, mutation rate $R$, perturbation size $\varepsilon$.}
\KwResult{perturbed image $x*$ (hopefully adversarial)}

\tcc{ Initialize population and fitness.}\
\For{$i \gets 1, \dots, N$}{ \
    $P_i \gets \text{random $\{-\varepsilon, \varepsilon\}$-valued vector}$\
}\

\For{$i \gets 1, \dots, N$}{
    $F_i \gets \fitness(P_i, c)$\
}

\tcc{Update population and fitness.}
\For{$g \gets 1, \dots, G$}{
    $P \gets \mutate(\cross(\select(P, F)), R)$\
    \For{$i \gets 1, \dots, N$}{
        $F_i \gets \fitness(P_i, c)$\
    }
}

\textbf{return} $x + P_{\argmax(F)}$

\caption{Genetic Attack}
\end{algorithm}

\begin{algorithm}[H] \footnotesize
\SetAlgoLined
\label{alg:select}
\KwData{Population $P$, fitnesses $F$.}
\KwResult{Array of parent perturbations.}
\For{$i \gets 1, \dots, \size(P)$}{
    select $j, k$ uniformly from $\{1, \dots, \size(P)\}$\
    \eIf{$F_j > F_k$}{
        $M_i \gets P_j$\
    }
    {
        $M_i \gets P_k$\
    }
}
\textbf{return} $M$
\caption{Select}
\end{algorithm}

\begin{algorithm}[H] \footnotesize
\SetAlgoLined
\label{alg:cross}
\KwData{Array $M$ of parent perturbations.}
\KwResult{Array of child perturbations.}
\For{$i \gets 2, 4, 6, \dots, \size(M)$}{

    $\vec{p} \gets P_{i-1}$ \ 
    
    $\vec{q} \gets P_i$ \ 
    
    select $t$ uniformly from $\{0, 1, \dots, \text{image side length}\}$ \ 
    
    select $u$ uniformly from $\{1, 2\}$ \
    \eIf{$u = 1$}{
        $\vec{c}, \vec{d} \gets$ cross $\vec{p}, \vec{q}$ horizontally at $t$ (Fig. \ref{fig:cross})\
    }
    {
        $\vec{c}, \vec{d} \gets$ cross $\vec{p}, \vec{q}$ vertically \
    }
    $C_{i-1} \gets \vec{c}$ \ 
    $C_i \gets \vec{d}$ \
}
\textbf{return} $C$\
\caption{Cross}
\end{algorithm}

\begin{algorithm}[H] \footnotesize
\SetAlgoLined
\label{alg:mutate}
\KwData{Array $C$ of child perturbations, mutation rate $R$.}
\KwResult{Array of mutated} 

\For{$i \gets 1, \dots, \size(C)$}{
    \If{$r \sim U(0, 1) < R$}{
        $\vec{c} \gets C_i$ \
        
        select pixel $j$ uniformly randomly \ 
        
        $\vec{c}_j \gets -\vec{c}_j$ \
    }
}
\textbf{return} $C$
\caption{Mutate}
\end{algorithm}

\end{document}